\documentclass[10pt,twocolumn,letterpaper]{article}

\usepackage{iccv}
\usepackage{times}
\usepackage{graphicx}
\usepackage{amsmath,amssymb,amsthm,mathabx}
\usepackage{booktabs}
\usepackage{acronym}
\usepackage{enumitem}
\usepackage[pagebackref=true,breaklinks=true,colorlinks,bookmarks=false]{hyperref}
\usepackage{balance}
\usepackage{setspace}
\usepackage[skip=3pt,font=small]{subcaption}
\usepackage[skip=3pt,font=small]{caption}
\usepackage[dvipsnames,table]{xcolor}
\usepackage[capitalise]{cleveref}
\usepackage{tabularx,multirow,array,makecell}
\usepackage{pifont}

\makeatletter
\renewcommand{\paragraph}{%
  \@startsection{paragraph}{4}%
  {\z@}{0ex \@plus 0ex \@minus 0ex}{-1em}%
  {\hskip\parindent\normalfont\normalsize\bfseries}%
}
\makeatother

\graphicspath{{figures/}}

\crefname{algocf}{Alg.}{Algs.}
\Crefname{algocf}{Algorithm}{Algorithm}

\definecolor{gblue}{HTML}{4285F4}
\definecolor{gred}{HTML}{DB4437}

\newcommand{\cmark}{\ding{51}}
\newcommand{\xmark}{\ding{55}}

\acrodef{ref}[REF]{Referring Expression Comprehension}
\acrodef{eru}[ERU]{Embodied Reference Understanding}
\acrodef{amt}[AMT]{Amazon Mechanic Turk}
\acrodef{roi}[RoIs]{Region of Interests}
\acrodef{iou}[IoU]{Intersection over Union}
\acrodef{paf}[PAF]{Part Affinity Field}

\usepackage[breaklinks=true,bookmarks=false]{hyperref}
\usepackage[accsupp]{axessibility}
\iccvfinalcopy 

\def\iccvPaperID{****} 

\ificcvfinal\fi

\begin{document}

\title{\vspace{-12pt}\textsl{YouRefIt}: Embodied Reference Understanding with Language and Gesture\vspace{-15pt}}

\author{Yixin Chen$^{1}$, Qing Li$^{1}$, Deqian Kong$^{1}$, Yik Lun Kei$^{1}$,\\
\vspace{3pt}Song-Chun Zhu$^{2,3,4}$, Tao Gao$^{1}$, Yixin Zhu$^{2,3}$, Siyuan Huang$^{1}$\\
$^{1}$ University of California, Los Angeles $^{2}$ Beijing Institute for General Artificial Intelligence\\
\vspace{3pt}$^{3}$ Peking University $^{4}$ Tsinghua University\\
\url{https://yixchen.github.io/YouRefIt}\vspace{-15pt}\\
}


\maketitle
\ificcvfinal\thispagestyle{empty}\fi

\setstretch{0.965}

\begin{abstract}
We study the machine's understanding of \textbf{embodied reference}: One agent uses both language and gesture to refer to an object to another agent in a shared physical environment. Of note, this new visual task requires understanding multimodal cues with perspective-taking to identify which object is being referred to. To tackle this problem, we introduce \textbf{\textsl{YouRefIt}}, a new crowd-sourced dataset of embodied reference collected in various physical scenes; the dataset contains 4,195 unique reference clips in 432 indoor scenes. To the best of our knowledge, this is the first embodied reference dataset that allows us to study referring expressions in daily physical scenes to understand referential behavior, human communication, and human-robot interaction. We further devise two benchmarks for image-based and video-based embodied reference understanding. Comprehensive baselines and extensive experiments provide the very first result of machine perception on how the referring expressions and gestures affect the embodied reference understanding. Our results provide essential evidence that gestural cues are as critical as language cues in understanding the embodied reference.
\end{abstract}

\section{Introduction}

\begin{figure}[t!]
	\centering
	\includegraphics[width=\linewidth]{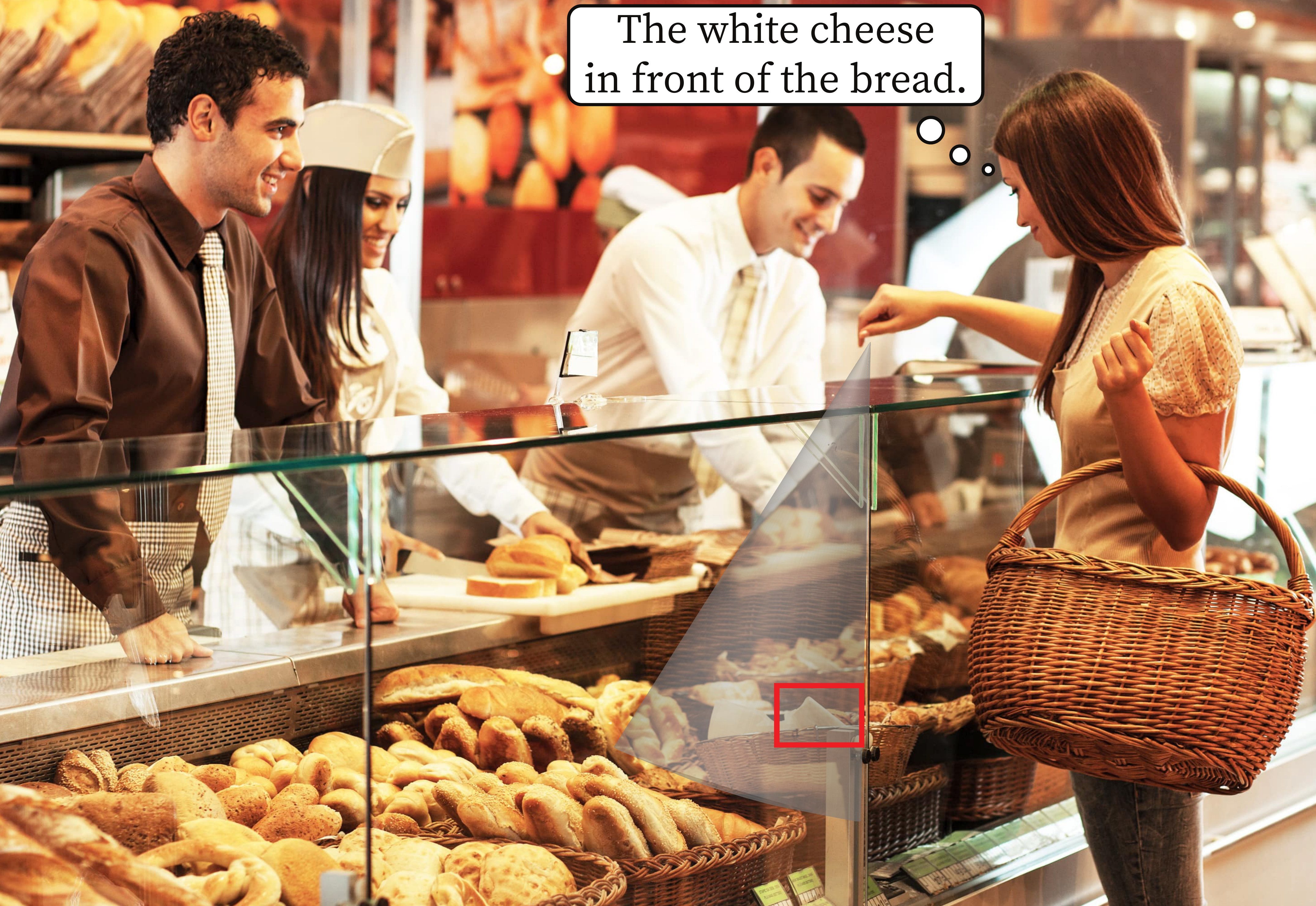}
	\caption{Imagine you walk into a bakery for your favorite. To precisely express your intent, you point to it and produce an utterance ``the white cheese in front of the bread.'' This multimodal communicative act is produced by assuming it can be properly understood by the staff, whose embodiment differs in the shared physical environment. Such a daily deictic-interaction scenario illustrates the significance of visual perspective-taking in embodied reference.}
	\label{fig:overview}
\end{figure}

Human communication~\cite{tomasello2010origins} relies heavily on establishing common ground~\cite{tang2020bootstrapping,stacy2020intuitive} by referring to objects in a shared environment. This process usually takes place in two forms: language (abstract symbolic code) and gesture (unconventionalized and uncoded). In the computer vision community, efforts of understanding reference have been primarily devoted in the first form through an artificial task, \ac{ref}~\cite{yu2016modeling,hu2017modeling,yu2018mattnet,liu2019improving,ye2019cross,yang2019cross,yang2020graph}, which localizes a particular object in an image with a natural language expression generated by the annotator. Evidently, the second form, gesture, has been left almost untouched. Yet, this nonverbal (gesture) form is more profound in the communication literature compared to the pure verbal (language) form with ample evolutionary evidence~\cite{arbib2008primate,mcneill2012language,halina2013ontogenetic}; it is deeply rooted in human cognition development~\cite{liszkowski2004twelve,liszkowski200612} and learning process~\cite{cook2008gesturing}, and tightly coupled with the language development~\cite{kita2003pointing,colonnesi2010relation,iverson2005gesture}.

Fundamentally, most modern literature deviates from the natural setting of reference understanding in daily scenes, which is \textbf{embodied}: An agent refers to an object to another in a \textit{shared} physical space~\cite{qiu2020human,wu2021communicative,fan2021learning}, as exemplified by \cref{fig:overview}. Embodied reference possesses two distinctive characteristics compared to \ac{ref}. First, it is \textbf{multimodal}. People often use both natural language and gestures when referring to an object. The gestural component and language component are semantically coherent and temporally synchronous to coordinate with one another, creating a concise and vivid message~\cite{kendon2004gesture} while elucidating the overloaded meaning if only one modality is presented~\cite{jiang2021individual}. Second, recognizing embodied reference requires visual \textbf{perspective-taking}~\cite{krauss1991perspective,batson1997perspective,qiu2020human}, the awareness that others see things from different viewpoints and the ability to imagine what others see from their perspectives. It requires both the message sender and receiver to comprehend the immediate environments~\cite{fan2021learning}, including the relationship between the interlocutors and the relationships between objects, in the shared perceptual fields for effective communication.

To address the deficiencies in prior work and study reference understanding at a full spectrum, we introduce a new dataset, \textbf{\textsl{YouRefIt}}, for embodied reference understanding. The reference instances in \textsl{YouRefIt} are crowd-sourced with diverse physical scenes from \ac{amt}. Participants are instructed to film videos in which they reference objects in a scene to an imagined person (\ie, a mounted camera) using both language and gestures. Minimum requirements of the scenes, objects, and words are imposed to ensure the naturalness and the variety of collected videos. Videos are segmented into short clips, with each clip containing an exact one reference instance. For each clip, we annotate the reference target (object) with a bounding box. We also identify \textbf{canonical frames} in a clip: They are the ``keyframes'' of the clip and contain sufficient information of the scene, human gestures, and referenced objects that can truthfully represent the reference instance. Fine-grained semantic parsing of the transcribed sentences is further annotated to support a detailed understanding of the sentences. In total, the \textsl{YouRefIt} dataset includes 4,195 embodied reference instances from 432 indoor scenes.

To measure the machine's ability in \acf{eru}, we devise two benchmarks on top of the proposed \textsl{YouRefIt} dataset. (i) \textbf{Image \ac{eru}} takes a canonical frame and the transcribed sentence of the reference instance as the inputs and predicts the bounding box of the referenced object. Image \ac{eru} adopts the settings from the well-studied \ac{ref} but is inherently more challenging and holistic due to its requirement on a joint and coherent understanding of human gestures, natural language, and objects in the context of human communication. (ii) \textbf{Video \ac{eru}} takes the video clip and the sentence as the input, identifies the canonical frames, and locates the reference target within the clip. Compared to Image \ac{eru}, Video \ac{eru} takes one step further and manifests the most natural human-robot communication process that requires distinguishing the initiation, the canonical frames, and the ending of a reference act while estimating the reference target in a temporal order.

Incorporating both language and gestural cues, we formulate a new multimodal framework to tackle the \ac{eru} tasks. In experiments, we provide multiple baselines and ablations. Our results reveal that models with explicit gestural cues yield better performance, validating our hypothesis that gestural cues are as critical as language cues in resolving ambiguities and overloaded semantics with cooperation (perspective-taking) in mind~\cite{jiang2021individual,jia2020lemma,qiu2020human,yuan2020joint,zhu2020dark}, echoing a recent finding in the embodied navigation task~\cite{wu2021communicative}. We further verify that temporal cues are essential in canonical frame detection, necessitating understanding embodied reference in dynamic and natural sequences.

This paper makes three major contributions. (i) We collect the first video dataset in physical scenes, \textsl{YouRefIt}, to study the reference understanding in an \textit{embodied} fashion. We argue this is a more natural setting than prior work and, therefore, further understanding human communications and multimodal behavior. (ii) We devise two benchmarks, Image \ac{eru} and Video \ac{eru}, as the protocols to study and evaluate the embodied reference understanding. (iii) We propose a multimodal framework for \ac{eru} tasks with multiple baselines and model variants. The experimental results confirm the significance of the joint understanding of language and gestures in embodied reference.

\section{Related Work}

Our work is related to two topics in modern literature: (i) \acf{ref} studied in the context of Vision and Language, and (ii) reference recognition in the field of Human-Robot Interaction. Below, we compare our work with prior arts on these two topics.

\begin{table*}[ht!]
    \caption{\textbf{Comparisons between the proposed \textsl{YouRefIt} and other reference datasets.} \textbf{Lang.} and \textbf{Gest.} denote whether language or gesture is used when referring to objects, and \textbf{Embo.} denotes whether referrers are embodied in the scenes where reference happens.}
    \label{tab:dataset_comparison}
    \resizebox{\textwidth}{!}{%
        \begin{tabular}{lcccccrrrr}
        \toprule
        \textbf{Datasets} &
        \textbf{Lang.} &
        \textbf{Gest.} &
        \textbf{Embo.} &
        \textbf{Type} &
        \textbf{Source} &
        \multicolumn{1}{c}{\begin{tabular}[c]{@{}c@{}}\textbf{No. of}\\\textbf{images} \end{tabular}} &
        \multicolumn{1}{c}{\begin{tabular}[c]{@{}c@{}}\textbf{No. of}\\\textbf{instances} \end{tabular}} &
        \multicolumn{1}{c}{\begin{tabular}[c]{@{}c@{}}\textbf{No. of object}\\\textbf{categories} \end{tabular}} &
        \multicolumn{1}{c}{\begin{tabular}[c]{@{}c@{}}\textbf{Ave. sent.}\\\textbf{length} \end{tabular}} \\
        \midrule
        PointAt~\cite{schauerte2010saliency}   & \xmark & \cmark & \cmark & image & lab         &    220   & 220     & 28  &    -   \\
        ReferAt~\cite{schauerte2010focusing}   & \cmark & \cmark & \cmark & video & lab         &    -   & 242      & 28  &    -   \\
        IPO~\cite{shukla2015probabilistic}     & \xmark & \cmark & \cmark & image & lab         &   278 & 278     & 10  &      - \\
        IMHF~\cite{shukla2016multi}            & \xmark & \cmark & \cmark & image & lab         &    1716   & 1,716   & - &    -   \\
        RefIt~\cite{kazemzadeh2014referitgame} & \cmark & \xmark & \xmark & image & image CLEF  & 19,894 & 130,525 & 238  & 3.61  \\
        RefCOCO~\cite{yu2016modeling}          & \cmark & \xmark & \xmark & image & MSCOCO      & 19,994 & 142,209 & 80  & 3.61  \\
        RefCOCO+~\cite{yu2016modeling}         & \cmark & \xmark & \xmark & image & MSCOCO      & 19,992 & 141,564 & 80  & 3.53  \\
        RefCOCOg~\cite{mao2016generation}         & \cmark & \xmark & \xmark & image & MSCOCO      & 26,711 & 104,560 & 80  & 8.43  \\
        Flickr30k entities~\cite{plummer2015flickr30k} & \cmark & \xmark & \xmark & image & Flickr30K   & 31,783 & 158,915 &  44,518   & -     \\
        GuessWhat?~\cite{de2017guesswhat}      & \cmark & \xmark & \xmark & image & MSCOCO      & 66,537 & 155,280 &  -   & -   \\
        Cops-Ref~\cite{chen2020cops}         & \cmark & \xmark & \xmark & image & COCO/Flickr & 75,299 & 148,712 & 508 & 14.40 \\
        CLEVR-Ref+~\cite{liu2019clevr}         & \cmark & \xmark & \xmark & image & CLEVR       & 99,992 & 998,743 & 3  & 22.40 \\
        \midrule
        \textsl{YouRefIt}                               & \cmark & \cmark & \cmark & video & crowd-sourced         & 497,348       & 4,195   & 395 & 3.73  \\ \bottomrule
        \end{tabular}%
    }%
\end{table*}

\subsection{\acf{ref}}

\ac{ref} is a visual grounding task. Given a natural language expression, it requires an algorithm to locate a particular object in a scene. Several datasets, including both images of physical scenes~\cite{kazemzadeh2014referitgame,yu2016modeling,mao2016generation,plummer2015flickr30k,de2017guesswhat,chen2020cops,cirik2020refer360,achlioptas2020referit3d} and synthetic images~\cite{liu2019clevr}, have been constructed by asking annotators or algorithms to provide utterances describing regions of images. To solve \ac{ref}, researchers have attempted various approaches~\cite{ye2019cross,liu2019improving,yang2019cross,yang2020graph}. Representative methods include (i) localizing a region by reconstructing the sentence using an attention mechanism~\cite{rohrbach2016grounding}, (ii) incorporating contextual information to ground referring expressions~\cite{zhang2018grounding,yu2016modeling}, (iii) using neural modular networks to better capture the structured semantics in sentences~\cite{hu2017modeling,yu2018mattnet}, and (iv) devising a one-stage approach~\cite{yang2019fast,yang2020improving}. In comparison, our work fundamentally differs from \ac{ref} at two levels.

\paragraph{Task-level}

\ac{ref} primarily focuses on building correspondence between visual cues and verbal cues (natural language). In comparison, the proposed \ac{eru} task mimics the minimal human communication process in an embodied manner, which requires a mutual understanding of both verbal and nonverbal messages signaled by the sender. Recognizing references in an embodied setting also introduces new challenges, such as visual perspective-taking~\cite{galinsky2008pays}: The referrers need to consider the perception from the counterpart's perspective for effective verbal and nonverbal communication, requiring a more holistic visual scene understanding both geometrically and semantically. In this paper, to study the reference understanding that echoes the above characteristics, we collect a new dataset containing natural reference scenarios with both language and gestures.

\paragraph{Model-level}

Since previous \ac{ref} approaches are only capable of comprehending communicative messages in the form of natural language and mostly ignore the gestural cues, it is insufficient in the \ac{eru} setting or to be applied in our newly collected dataset. To tackle this deficiency, we design a principled framework to combine verbal (natural language) and nonverbal (gestures) cues. The proposed framework outperforms prior single-modality methods, validating the significant role of the gestural cue in addition to the language cue in embodied reference understanding.

\setstretch{1}

\subsection{Reference in Human-Robot Interaction}

The combination of verbal and nonverbal communication for reference is one of the central topics in Human-Robot Interaction. Compared with \ac{ref}, this line of work focuses on more natural settings but with specialized scenarios. One stream of work emphasizes pointing direction and thus are not object-centric while missing language reference: The Innsbruck Pointing at Objects dataset~\cite{shukla2015probabilistic} investigates two types of pointing gestures with index finger and tool, and the Innsbruck Multi-View Hand Gesture Dataset~\cite{shukla2016multi} records hand gestures in the context of human-robot interaction in close proximity. The most relevant prior arts are ReferAt~\cite{schauerte2010focusing} and PointAt~\cite{schauerte2010saliency}, wherein participants are tasked to point at various objects with or without linguistic utterance. Some other notable literature includes (i) a robotics system that allows users to combine natural language and pointing gestures to refer to objects on a display~\cite{kobsa1986combining}, (ii) experiments that investigate the semantics and pragmatics of co-verbal pointing through computer simulation~\cite{lucking2015pointing}, (iii) deictic interaction with a robot when referring to a region using pointing and spatial deixis~\cite{hato2010pointing}, and (iv) effects of various referential strategies, including talk-gesture-coordination and handshape, for robots interacting with humans when guiding attentions in museums~\cite{pitsch2014robot}.

Although related, the above literature is constrained in lab settings with limited sizes, scenarios, and expressions, thus insufficient for solving the reference understanding in natural, physical scenarios with both vision and language. In comparison, crowd-sourced by \ac{amt}, our dataset is much more diverse in environment setting, scene appearance, and types of utterance. Our dataset also collects videos instead of static images commonly used in prior datasets, opening new venues to study dynamic and evolutionary patterns that occurred during natural human communications.

\section{The \textsl{YouRefIt} Dataset}

To study the embodied reference understanding, we introduce a new dataset named \textsl{YouRefIt}, a video collection of people referring to objects with both natural language and gesture in indoor scenes. \cref{tab:dataset_comparison} tabulates a detailed comparison between \textsl{YouRefIt} against twelve existing reference understanding datasets. Compared to existing datasets collected either in laboratories or from the Internet (MSCOCO/Flickr) or simulators (CLEVR), \textsl{YouRefIt} has a clear distinction: It contains videos crowd-sourced by \ac{amt}, and thus the reference happens in a more natural setting with richer diversity. Compared with the datasets on referring expression comprehension, the referrers (human) and the receivers (camera) in our dataset share the same physical environment, with both language and gesture allowed for referring to objects; the algorithm ought to understand from an embodiment perspective to tackle this problem. Next, we discuss the data collection and annotation process details, followed by a comprehensive analysis.

\begin{figure}[t!]
	\centering
	\includegraphics[width=\linewidth]{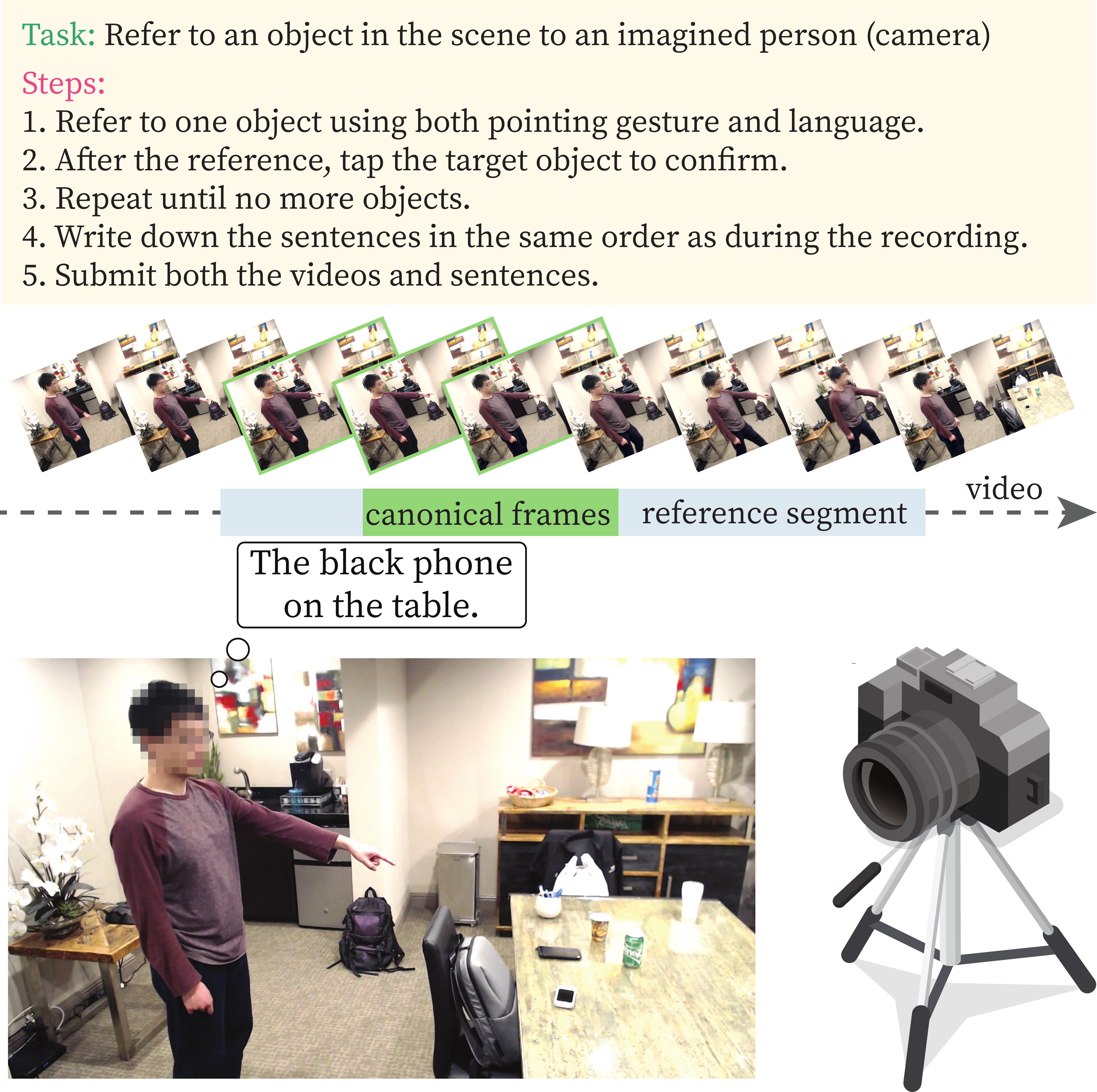}
	\caption{\textbf{Illustration of the dataset collection procedure.} Participants were asked to film a series of reference tasks to an imaged person (\ie, the camera) following the instructions.}
	\label{fig:data_collection}
\end{figure}

\subsection{Data Collection}

Our dataset was collected via \ac{amt}; see the data collection process in \cref{fig:data_collection}. Workers were asked to record a video containing actions of referring to objects in the scene to an imagined person (\ie, the camera) using both natural languages (sentences) and pointing gestures. Most videos were collected in indoor scenes, such as offices, kitchens, and living rooms. Unlike existing datasets in which objects are usually put on a table with a clean background, all the objects in our collected videos were placed at their natural positions. Each video also included more than ten objects in the scene to avoid trivial scenarios and increase the reference difficulty. The camera was set up such that the referrer and all referred objects are within the field of view.

When referring to a specific object, participants were instructed to use arbitrary natural languages and gestures freely. However, they were also required to avoid potential ambiguities, such that the observer would be able to uniquely identify the referred object by merely observing the reference behaviors. After reference actions were finished, participants were instructed to tap the referred object; this extra step helps annotate the referred target. In addition to the voices recorded in the video, participants were also asked to write down the sentences after the recording.

\subsection{Data Annotation}

The annotation process takes two stages: (i) annotation of temporal segments, canonical frames, and referent bounding boxes, and (ii) annotation of sentence parsing. Please refer to the \textit{supplementary material} for more details of the data post-processing and annotation process.

\paragraph{Segments}

Since each collected video consists of multiple reference actions, we first segment the video into clips; each contains an exact one reference action. A segment is defined from the start of gesture movement or utterance to the end of the reference, which typically includes the raise of hand and arm, pointing action, and reset process, synchronized with its corresponding language description.

\paragraph{Canonical Frames}

In each segment, the annotators were asked to annotate further the canonical moments, which contain the ``keyframes'' that the referrer holds the steady pose to indicate what is being referred clearly. Combined with natural language, it is sufficient to use any canonical frame to localize the referred target.

\paragraph{Bounding Boxes}

Recall that participants were instructed to tap the referred objects after each reference action. Using this information, bounding boxes of the referred objects were annotated using Vatic~\cite{vondrick2013efficiently}, and the tapping actions were discarded. The object color and material were also annotated if identifiable. The taxonomy of object color and material is adopted from Visual Genome dataset~\cite{krishna2017visual}.

\paragraph{Sentence Parsing}

Given the sentence provided by the participants who performed reference actions, \ac{amt} annotators were asked to refine the sentence further and ensure it matches the raw audio collected from the video. We further provided more fine-grained parsing results of the sentence for natural language understanding. \ac{amt} annotators annotated target, target-attribute, spatial-relation, and comparative-relation. Take ``The largest red bottle on the table'' as an example: ``the bottle'' will be annotated as the target, ``red" as target-attribute, ``on the table'' as spatial-relation, and ``largest'' as comparative-relation. For each relation, we further divided them into ``relation'' (\eg, ``on'') and ``relation-target'' (\eg, ``the table'').

\subsection{Dataset Statistics}

In total, \textsl{YouRefIt} includes 432 recorded videos and 4,195 localized reference clips with 395 object categories. We retrieved 8.83 hours of video during the post-processing and annotated 497,348 frames. The total duration of all the reference actions is 3.35 hours, with an average duration of 2.81 seconds per reference. Each reference process was annotated with segments, canonical frames, bounding boxes of the referred objects, and sentences with semantic parsing. All videos were collected with synchronized audio. We also included the body poses and hand keypoints of the participants extracted by the OpenPose~\cite{cao2019openpose}.

\begin{figure*}[t!]
\centering
	\begin{subfigure}[t]{0.37\linewidth}
		\includegraphics[width=\linewidth,trim={0.5cm 0.8cm 0.4cm 0.5cm},clip]{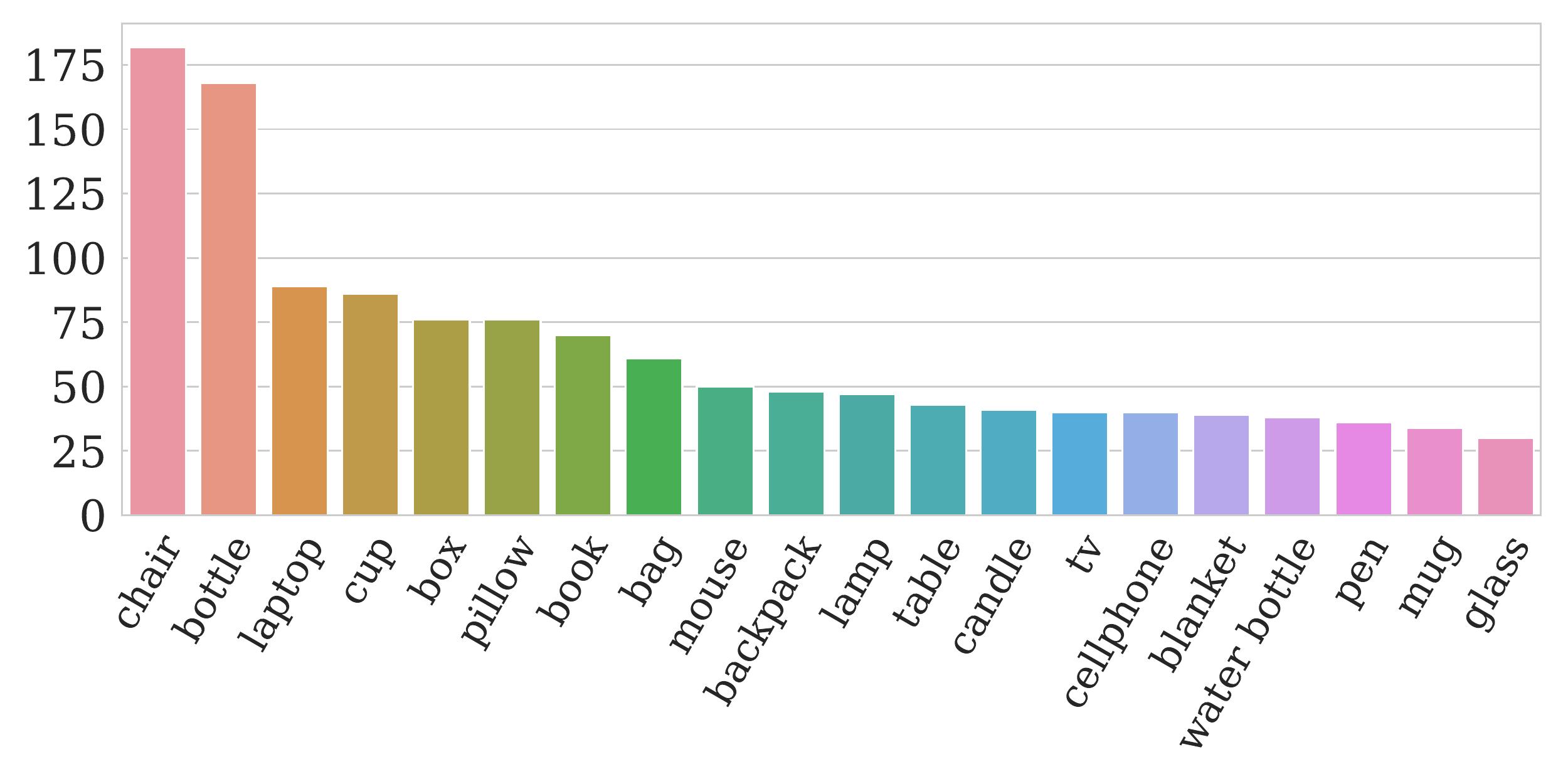}
    	\caption{The frequency of the top-20 referred objects.}
    	\label{fig:object_freq}
	\end{subfigure}%
	\begin{subfigure}[t]{0.35\linewidth}
		\includegraphics[width=\linewidth,trim={0cm 0cm 0cm 0cm},clip]{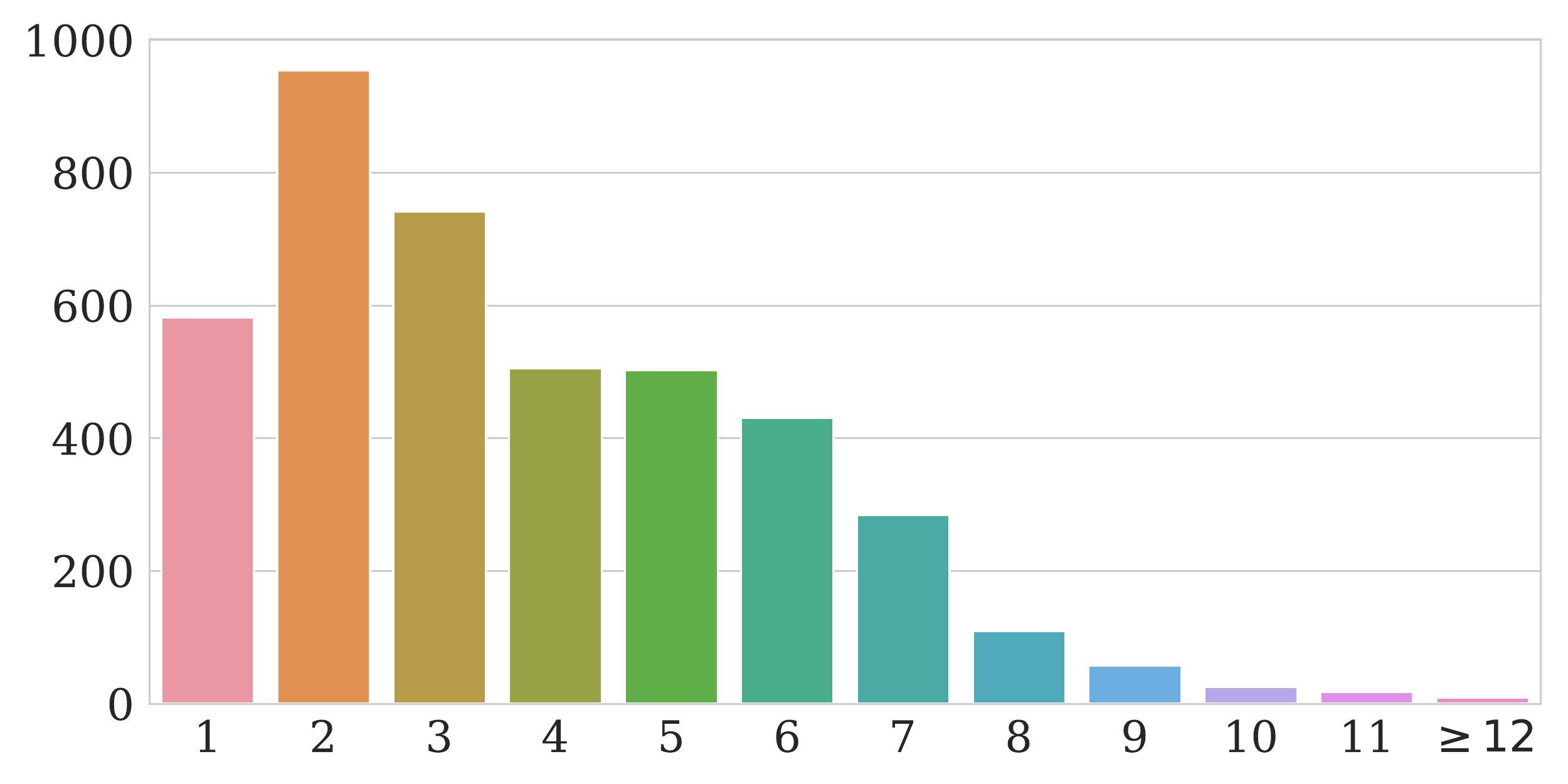}
    	\caption{The distribution of sentence lengths.}
    	\label{fig:sent_len}
	\end{subfigure}%
	\begin{subfigure}[t]{0.27\linewidth}
		\includegraphics[width=\linewidth,trim={0cm 0cm 0cm 0cm},clip]{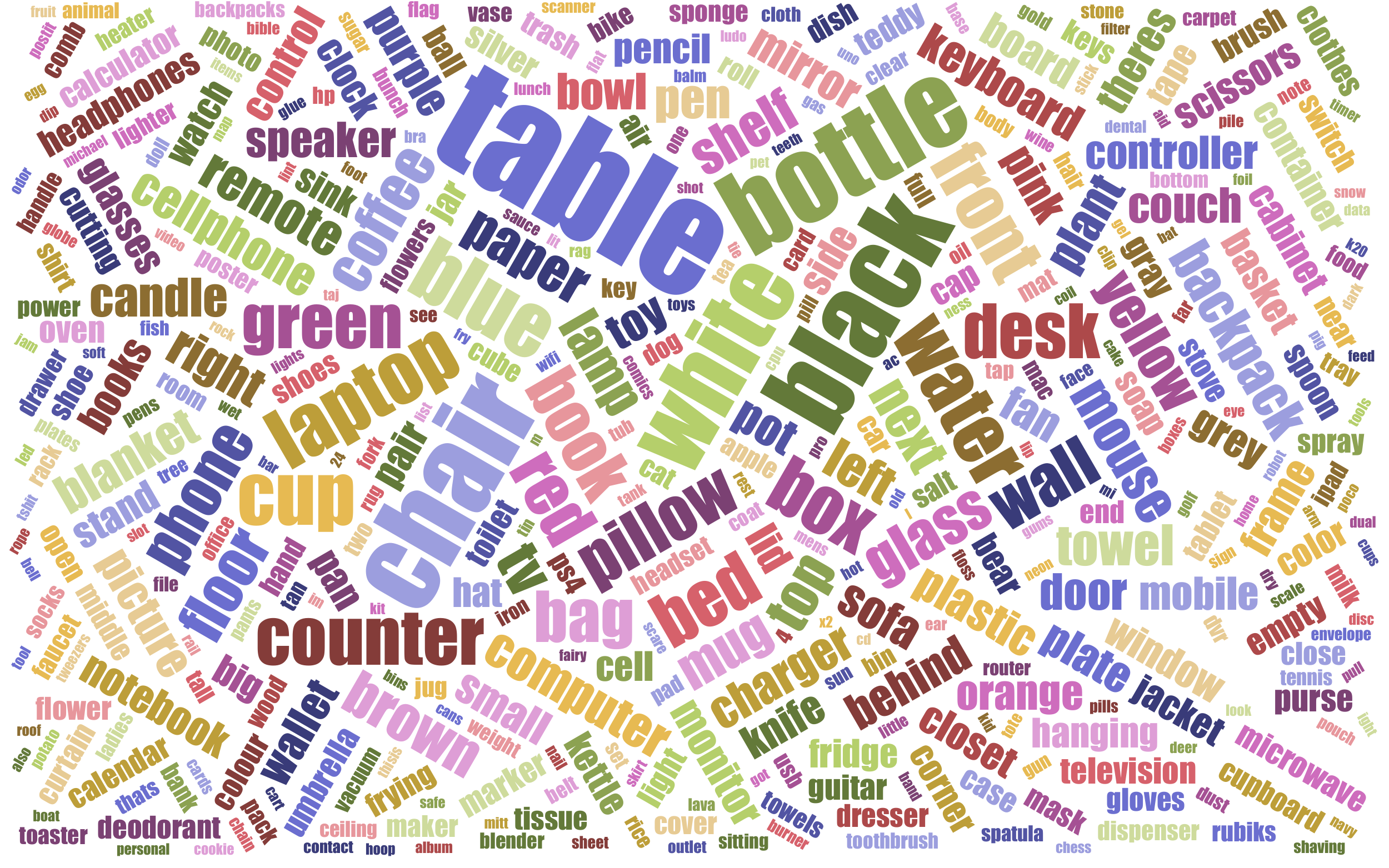}
    	\caption{The language wordle.}
    	\label{fig:word_cloud}
	\end{subfigure}%
    \caption{Statistics of the \textsl{YouRefIt} dataset.}
    \label{fig:statistics}
\end{figure*}

\paragraph{Object Categories}

\cref{fig:object_freq} shows the frequencies of the top-20 referred object categories, which roughly follow the Zipf's law~\cite{zipf1949human}. Since most videos were shot in indoor scenes, the most frequently referred are daily objects, such as ``chair,'' ``bottle,'' and ``cup.''

\paragraph{Reference Sentence}

\cref{fig:word_cloud} shows the word cloud of sentences after removing the stop words. Interestingly, the most frequent word is ``table,'' which is not even in the top-5 referred objects. A further inspection implies that the ``table'' is the most frequently used relational object while referring to objects by natural languages. \cref{fig:sent_len} shows the distribution of sentence lengths with an average of 3.73. We observe that the sentences in \textsl{YouRefIt} are much shorter than those of language-only reference datasets (\eg, 8.43 for RefCOCOg and 14.4 for Cops-Ref). This discrepancy implies that while naturally referring to objects, humans prefer a multimodal communication pattern that combines gestures with fewer words (compared to using a single modality) to minimize the cognitive load~\cite{sweller1998cognitive}.

\setstretch{0.98}

\section{\acf{eru}}

In this section, we benchmark two tasks of embodied reference understanding on the \textsl{YouRefIt} dataset, namely, Image \ac{eru} and Video \ac{eru}. The first benchmark evaluates the performance of understanding embodied reference based on the canonical frame, whereas the second benchmark emphasizes how to effectively recognize the canonical moments and reference targets simultaneously in a video sequence. Below, we describe the detailed settings, baselines, analyses, and ablative studies in the experiments.

\paragraph{Dataset Splits}

We randomly split the dataset into the training and test sets with a ratio of 7:3, resulting in 2,950 instances for training and 1,245 instances for testing.

\subsection{Image \ac{eru}}

Given the canonical frame and the sentence from an embodied reference instance, Image \ac{eru} aims at locating the referred object in the image through both the human language and gestural cues.

\paragraph{Experimental Setup and Evaluation Protocol}

For each reference instance, we randomly pick one frame from the annotated canonical frames. We adopt the evaluation protocol similar to the one presented in Mao \etal~\cite{mao2016generation}: (i) predict the region referred by the given image and sentence, (ii) compute the \ac{iou} ratio between the ground-truth and the predicted bounding box, and (iii) count it as correct if the \ac{iou} is larger; otherwise wrong. We use accuracy as the evaluation metric. Following object detection benchmark~\cite{geiger2012are}, we report the results under three different \ac{iou}s: 0.25, 0.5, and 0.75.

We also evaluate on subsets with various object sizes, \ie, \textit{small}, \textit{medium} and \textit{large}. Object size is estimated using the ratio between the area of the ground-truth object bounding box and the area of the image. The size thresholds are $0.48\%$ and $1.76\%$ based on the size distribution in the dataset; see the size distribution in \textit{supplementary material}.

\begin{figure*}[t!]
	\centering
	\includegraphics[width=\linewidth]{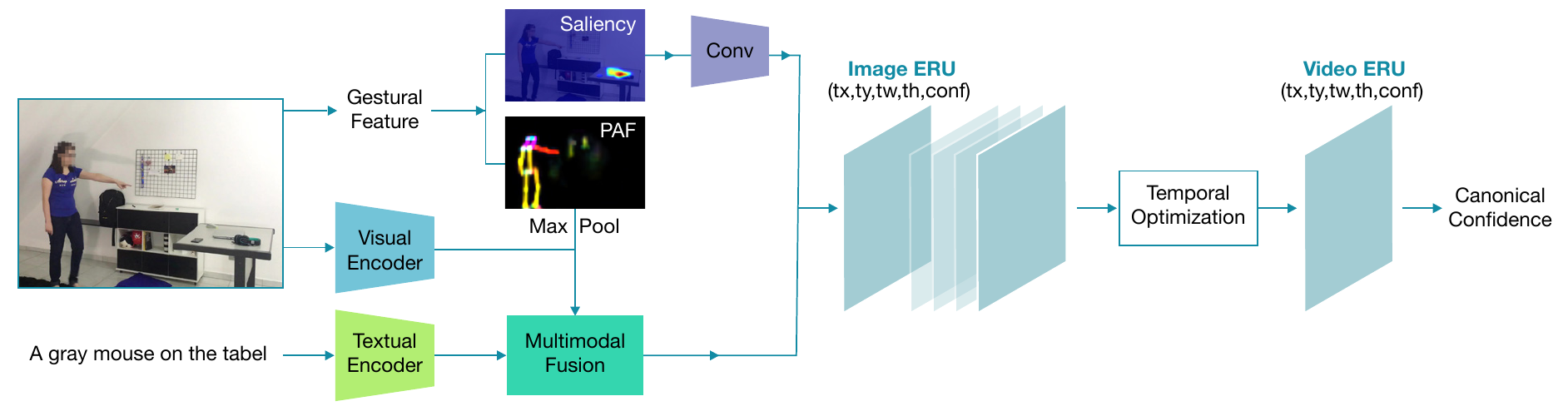}
	\caption{The proposed multimodal framework for the \ac{eru} task that incorporates both human gestural cues and language cues.}
	\label{fig:framework}
\end{figure*}

\paragraph{Methods}

We devise a novel multimodal framework for Image \ac{eru} that leverages both the language and gestural cues; see \cref{fig:framework}. At a high level, our framework includes both the visual and language encoder, similar to prior \ac{ref} models~\cite{yang2019fast,yang2020improving,luo2020multi}, as well as explicitly extracted gesture features. We utilize the features from three modalities to effectively predict the target bounding box.

\setstretch{1}

\begin{figure*}[t!]
	\begin{subfigure}[t]{\linewidth}
		\includegraphics[width=\linewidth,trim={0cm 0cm 0cm 0cm},clip]{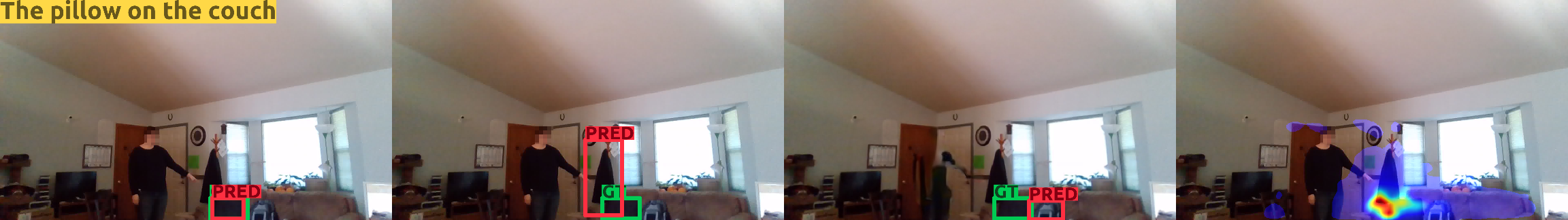}
	\end{subfigure}%
	\\
	\begin{subfigure}[t]{\linewidth}
		\includegraphics[width=\linewidth,trim={0cm 0cm 0cm 0cm},clip]{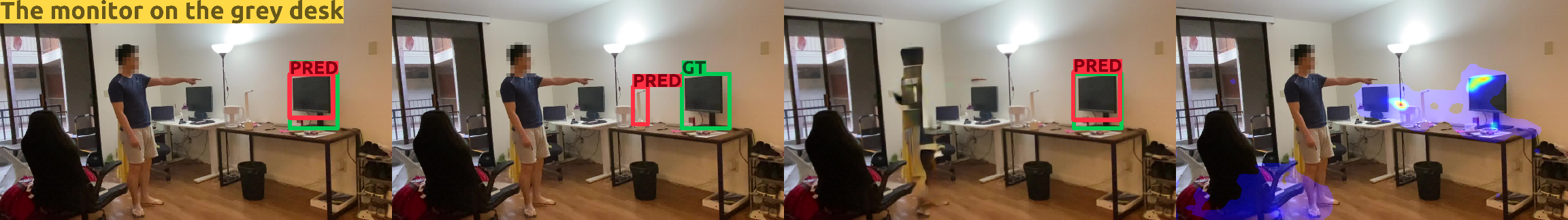}
	\end{subfigure}%
	\\
	\begin{subfigure}[t]{\linewidth}
		\includegraphics[width=\linewidth,trim={0cm 0cm 0cm 0cm},clip]{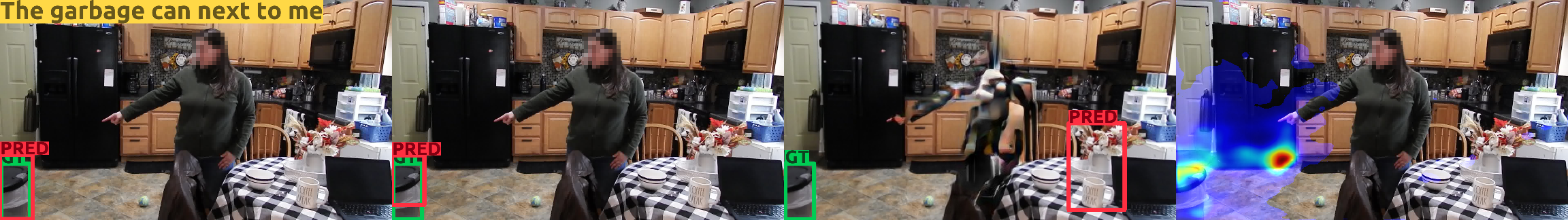}
	\end{subfigure}%
	\\
	\hspace*{1.3cm} (a) Ours$_{\text{Full}}$ \hspace{2.2cm} (b) Ours$_{\text{no\_lang}}$ \hspace{2.05cm} (c) ReSC$_{\text{inpaint}}$ \hfill (d) Saliency Map \hspace{0.95cm}
    \caption{\textbf{Qualitative results in Image \ac{eru} of representative models with various information sources and pointing saliency map.} Green/red boxes are the predicted/ground-truth reference targets. Sentences used during the references are shown at the top-left corner.}
    \label{fig:img_eru_qual}
\end{figure*}

Specifically, we use Darknet-53~\cite{redmon2018yolov3} pre-trained on COCO object detection~\cite{lin2014microsoft} as the visual encoder. The textual encoder is the uncased base version of BERT~\cite{devlin2018bert} followed by two fully connected layers. We incorporate two types of gestural features: (i) the \ac{paf}~\cite{cao2019openpose} heatmap, and (ii) the pointing saliency heatmap. Inspired by the visual saliency prediction, we train MSI-Net~\cite{kroner2020contextual} on the \textsl{YouRefIt} dataset to predict the salient regions by considering both the latent scene structure and the gestural cues, generating more accurate guidance compared to the commonly used \ac{roi}; see some examples of predicted salient regions in \cref{fig:img_eru_qual}. We aggregate the visual feature and \ac{paf} heatmaps by max-pooling and concatenation, fusing them with textual features by updating text-conditional visual features attended to different words through a sub-query module~\cite{yang2020improving}. Following convolution blocks, the saliency map feature is concatenated with the text-conditional visual feature as the high-level guidance to predict anchor boxes and confidence scores; we use the same classification and regression loss as in Yang \etal~\cite{yang2019fast} for anchor-based bounding box prediction.

\paragraph{Baselines and Ablations}

We first evaluate the Image \ac{eru} performance on FAOA~\cite{yang2019fast} and ReSC~\cite{yang2020improving}, originally designed for the \ac{ref} task. We also design baselines to test the gestural cues in a two-stage architecture, similar to MAttNet~\cite{yu2018mattnet}. We generate the \ac{roi}s by Region Proposal Network from Faster R-CNN~\cite{ren2016faster} pre-trained on the MSCOCO dataset. To score the object proposal, we test two categories of heatmaps that reflect the gestural cues. (i) By pointing heatmap from the primary pointing direction characterized by arm, hand, and index finger. Following Fan \etal~\cite{fan2018inferring}, we generate the pointing heatmap by a Gaussian distribution to model the variation of a pointing ray \wrt the primary pointing direction. We choose $15^{\circ}$ and $30^{\circ}$ as the standard deviations (\ie, RPN$_{\text{pointing15}}$ and RPN$_{\text{pointing30}}$). (ii) By pointing saliency map (\ie, RPN$_{\text{saliency}}$). The scores are computed according to the heatmap of average density.

We design ablation studies from two aspects: data and architecture. For the \textbf{data-wise} ablation, we first evaluate the MattNet, FAOA, and ReSC models pre-trained on the \ac{ref} datasets RefCOCO, RefCOCO+, and RefCOCOg, where the references are not embodied. Therefore, these three pre-trained models neglect the human gestural cues. Next, for a fair comparison without the gestural cues, we further generate an inpainted version of \textsl{YouRefIt}, where humans are segmented and masked by a pre-trained Mask R-CNN~\cite{he2017mask,wu2019detectron2}, and the masked images are inpainted by DeepFill~\cite{yu2018generative,yu2018free} pre-trained on the Places2~\cite{zhou2017places} dataset; see examples in \cref{fig:img_eru_qual}. After the human gestural cues are masked out, we train FAOA and ReSC on the inpainted dataset, denoted as FAOA$_{\text{inpaint}}$ and ReSC$_{\text{inpaint}}$. For the \textbf{architecture-wise} ablation, we compare two variants of our proposed full model to evaluate the contribution of different components: (i) Ours$_{\text{no\_lang}}$: without the language embedding module, and (ii) Ours$_{\text{PAF\_only}}$: with the \ac{paf} heatmap as the only gestural cue; see the \textit{supplementary material} for more details.

\paragraph{Results and Discussion}

\cref{tab:image_eru} tabulates the quantitative results of the Image \ac{eru}, and \cref{fig:img_eru_qual} shows some qualitative results. We categorize the models based on their information sources: \textit{Language-only}, \textit{Gesture-only}, and \textit{Language + Gesture}. Below, we summarize some key findings.
\begin{enumerate}[leftmargin=*,noitemsep,nolistsep]
    \item Gestural cues are essential for embodied reference understanding. As shown in \cref{tab:image_eru}, FAOA and ReSC models show significant performance improvement when trained on the original \textsl{YouRefIt} dataset compared to that on the inpainted version. Of note, in embodied reference, the referrer will adjust their own position to ensure the referred targets are not blocked by its body, one of the main advantages introduced by perspective-taking. As such, the inpainted images always contain the reference targets with only gestural cues masked.
    
    \item Language cues elucidate ambiguities where the gestural cues alone cannot resolve. As shown by the \textit{Gesture-only} models, RPN$+$heatmap models possess ambiguities when presented with gestural cues alone; pointing gestures suppress the descriptions of target location and attend to spatial regions but are not object-centric. Without the referring expressions, the performance of Ours$_{\text{no\_lang}}$ also deteriorates compared to Ours$_{\text{Full}}$.
    
    \item Explicit gestural features are beneficial for understanding embodied reference. Ours$_{\text{PAF\_only}}$, which incorporates \ac{paf} features that encode unstructured pairwise relationships between body parts, outperforms the original FAOA and ReSC models. By further adding the saliency heatmap, our full model Ours$_{\text{Full}}$ achieves the best performance in all baselines and ablations. Taken together, these results strongly indicate that the fusion of the language and gestural cues could be the crucial ingredient to achieving high model performance.
\end{enumerate}

\begin{table*}[ht!]
    \centering
    \caption{Comparisons of Image \ac{eru} performances on the \textsl{YouRefIt} dataset.}
    \label{tab:image_eru}
    \resizebox{\hsize}{!}{%
    \begin{tabular}{lcccccccccccc}
        \toprule
        \multirow{2}{*}{\textbf{Model}} & 
        \multicolumn{4}{c}{\shortstack[c]{\textbf{\ac{iou}=0.25}}} & 
        \multicolumn{4}{c}{\shortstack[c]{\textbf{\ac{iou}=0.5}}} & 
        \multicolumn{4}{c}{\shortstack[c]{\textbf{\ac{iou}=0.75}}} 
        \\
        \cline{2-13}
        &\textit{all} & \textit{small} & \textit{medium} & \textit{large} &\textit{all} &\textit{small} & \textit{medium} & \textit{large} & \textit{all}  & \textit{small} & \textit{medium} & \textit{large} 
        \\
        \hline
        \rowcolor[rgb]{ .682,  .667,  .667} \multicolumn{1}{l}{\textbf{Language-only}} & \multicolumn{12}{l}{}\\
        MAttNet$_{\text{pretrain}}$ &14.2&2.3&4.1&34.7&12.2&2.4&3.8&29.2&9.1&1.0&2.2&23.1\\
        FAOA$_{\text{pretrain}}$ & 15.9 &2.1 &9.5 &34.4 &11.7 &1.0 &5.4 &27.3 &5.1 &0.0 &0.0 &14.1  \\
        FAOA$_{\text{inpaint}}$ & 23.4 & 14.2 & 23.6 & 32.1 &16.4 &9.0 &17.9 &22.5 &4.1 &1.4 &4.7 &6.2  \\
        ReSC$_{\text{pretrain}}$
        &20.8&3.5&17.5&40.0&16.3&0.5&14.8&36.7&7.6&0.0&4.3&17.5\\
        ReSC$_{\text{inpaint}}$ &34.3&20.3&38.9&44.0&25.7&8.1&32.4&36.5&9.1&1.1&10.1&16.0\\
        
        \rowcolor[rgb]{ .682,  .667,  .667}
        \multicolumn{1}{l}{\textbf{Gesture-only}}& \multicolumn{12}{l}{}\\
        RPN+Pointing$_{15}$ &15.3&10.5&16.9&18.3&10.2&7.2&12.4&11.0&6.5&3.8&9.1&6.6\\
        RPN+Pointing$_{30}$ &14.7 &10.8&17.0&16.4&9.8&7.4&12.4&9.8&6.5&3.8&8.9&6.8 \\
        RPN+Saliency\cite{kroner2020contextual} &27.9&29.4&34.7&20.3&20.1&\textbf{21.1}&26.8&13.2&12.2&\textbf{10.3}&\textbf{17.9}&8.6\\
        Ours$_{\text{no\_lang}}$ &41.4&29.9&48.3&46.3&30.6&17.4&37.0&37.4&10.8&1.7&13.9&16.6\\
        \rowcolor[rgb]{ .682,  .667,  .667}
        \multicolumn{1}{l}{\textbf{Language + Gesture}}& \multicolumn{12}{l}{}\\
        FAOA\cite{yang2019fast} &44.5 &30.6&48.6&54.1&30.4&15.8&36.2&39.3&8.5 &1.4&9.6&14.4 \\
        ReSC\cite{yang2020improving} &49.2&32.3&54.7&60.1&34.9&14.1&42.5&47.7&10.5&0.2&10.6&20.1\\
        Ours$_{\text{PAF\_only}}$ &52.6&35.9&60.5&61.4&37.6&14.6&49.1&49.1&12.7&1.0&16.5&20.5\\
        Ours$_{\text{Full}}$ &\textbf{54.7}&\textbf{38.5}&\textbf{64.1}&\textbf{61.6}&\textbf{40.5} &16.3&\textbf{54.4}&\textbf{51.1}&\textbf{14.0} &1.2&17.2&\textbf{23.3}\\
        \midrule
        \rowcolor[rgb]{ .682,  .667,  .667} \textbf{Human}& 94.2$\pm$0.2 & 93.7$\pm$0.0 & 92.3$\pm$1.3 & 96.3$\pm$1.7 & 85.8$\pm$1.4 & 81.0$\pm$2.2 & 86.7$\pm$1.9 & 89.4$\pm$1.7 & 53.3$\pm$4.9 & 33.9$\pm$7.1 & 55.9$\pm$6.4 & 68.1$\pm$3.0 \\
        \bottomrule
    \end{tabular}
    }
\end{table*}

\paragraph{Human Performance}

We also conducted a human study of the embodied reference understanding task. We ask three Amazon Turkers to annotate the referred object bounding box in 1,000 images randomly sampled from the test set. We report the average accuracy under different \ac{iou}s in \cref{tab:image_eru}. Humans achieve significantly higher accuracy than all current machine learning models, demonstrating the human's outstanding capability to understand embodied references combined with language and gestural cues. The performance drops when the \ac{iou} threshold increases, especially for \textit{small} and \textit{medium} objects, indicating the difficulties in resolving the ambiguity in small objects.

\setstretch{0.98}

\subsection{Video \ac{eru}}

Compared with Image \ac{eru} discussed above, Video \ac{eru} is a more natural and practical setting in human-robot interaction. Given a referring expression and a video clip that captures the whole dynamics of a reference action with consecutive body movement, Video \ac{eru} aims at recognizing the canonical frames and estimate the referred target at the same time.

\paragraph{Experimental Setup and Evaluation Protocol}

For each reference instance, we sample image frames with 5 FPS from the original video clip. Average precision, recall, and F1-score are reported for the canonical frame detection. For referred bounding box prediction, we report the averaged accuracy in all canonical frames.

\paragraph{Baselines}

To further exploit the temporal constraints in videos, we integrate a temporal optimization module to aggregate and optimize the multimodal feature extracted from the Image \ac{eru}. We test two designs of temporal optimization module: (i) ConvLSTM: a two-layer convolutional Long Short-Term Memory~\cite{shi2015convolutional}, and (ii) Transformer: a three-layer Transformer encoder~\cite{vaswani2017attention} with four attention heads in each layer. After the temporal optimization module, we use the features of each frame to predict canonical frames and anchor bounding boxes simultaneously. 

We further design a third \textit{Frame-based} baseline that learns from the individual frame by adding two fully connected regression layers on top of our model in Image \ac{eru}. This \textit{Frame-based} model takes all sampled frames from the video clip during training and testing.

During training, we add a binary cross-entropy loss for canonical frame detection on top of the loss function for bounding box prediction in the Image \ac{eru} framework. Please refer to the \textit{supplementary material} for more details.

\begin{table*}[ht!]
    \centering
    \caption{Video \ac{eru} performance comparisons on the \textsl{YouRefIt} dataset.}
    \begin{tabular}{lcccccccccccc}
        \toprule
        \multirow{2}{*}{\textbf{Model}} & 
        \multicolumn{4}{c}{\shortstack[c]{\textbf{\ac{iou}=0.25}}} & 
        \multicolumn{4}{c}{\shortstack[c]{\textbf{\ac{iou}=0.5}}} & 
        \multicolumn{4}{c}{\shortstack[c]{\textbf{\ac{iou}=0.75}}} 
        \\
        &\textit{all} & \textit{small} & \textit{medium} & \textit{large} &\textit{all} &\textit{small} & \textit{medium} & \textit{large} & \textit{all}  & \textit{small} & \textit{medium} & \textit{large} 
        \\
        \midrule
        Frame-based & \textbf{55.2} &42.3&\textbf{58.9}  &\textbf{64.8} &\textbf{41.7} &\textbf{22.7} &53.4 &\textbf{48.8} &16.9 &1.6 &21.8 &\textbf{27.0}  \\
        Transformer & 52.3 &40.2&55.6&58.3&38.8&21.2&54.1 &47.1&13.9 &1.5&20.8 &22.7  \\
        ConvLSTM
        &54.8&\textbf{43.1}&57.5&60.0&39.3&22.5&\textbf{54.8}&46.7&\textbf{17.3}&\textbf{1.8}&\textbf{24.3}&25.5\\
        \midrule
        Ours$_{\text{Full}}$ &54.7&38.5&64.1&61.6&40.5 &16.3&54.4&51.1&14.0 &1.2&17.2&23.3\\
        \bottomrule
    \end{tabular}
    \label{tab:video_eru}
\end{table*}

\begin{figure*}[t!]
    \begin{subfigure}[t]{0.25\linewidth}
    	\includegraphics[width=\linewidth,trim={0cm 0cm 0cm 0cm},clip]{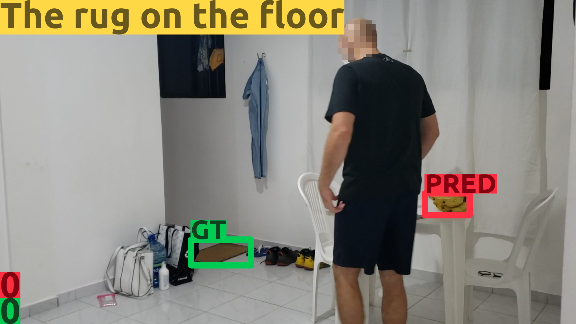}
    \end{subfigure}%
    \begin{subfigure}[t]{0.25\linewidth}
    	\includegraphics[width=\linewidth,trim={0cm 0cm 0cm 0cm},clip]{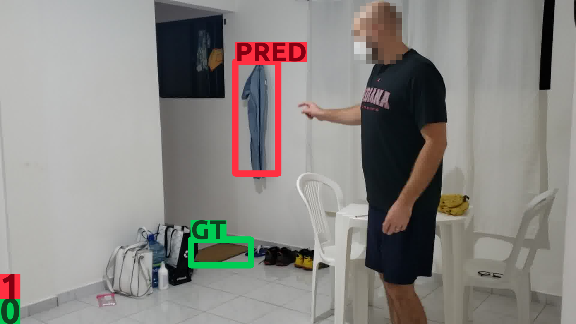}
    \end{subfigure}%
    \begin{subfigure}[t]{0.25\linewidth}
    	\includegraphics[width=\linewidth,trim={0cm 0cm 0cm 0cm},clip]{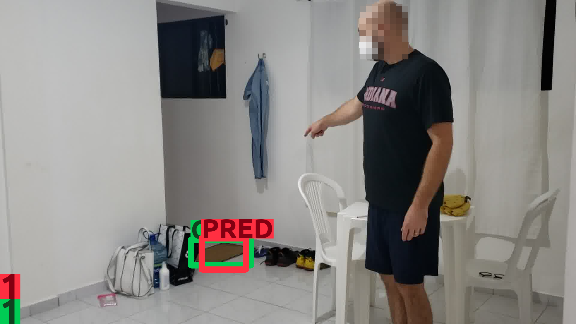}
    \end{subfigure}%
    \begin{subfigure}[t]{0.25\linewidth}
    	\includegraphics[width=\linewidth,trim={0cm 0cm 0cm 0cm},clip]{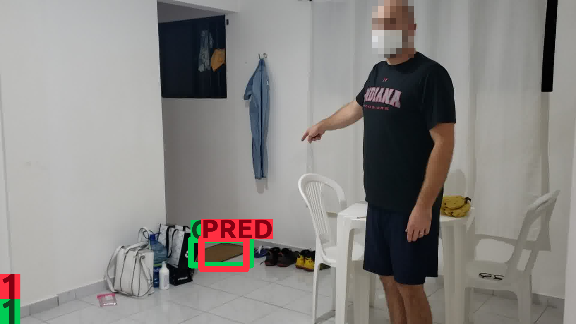}
    \end{subfigure}%
    \\
    \begin{subfigure}[t]{0.25\linewidth}
    	\includegraphics[width=\linewidth,trim={0cm 0cm 0cm 0cm},clip]{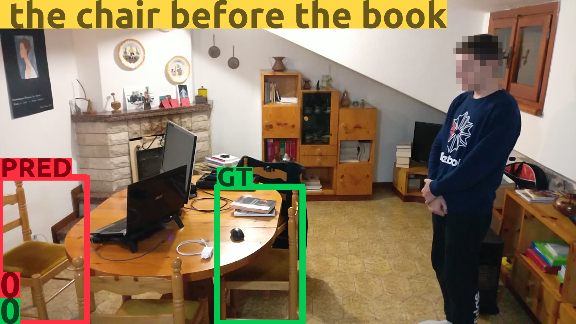}
    \end{subfigure}%
    \begin{subfigure}[t]{0.25\linewidth}
    	\includegraphics[width=\linewidth,trim={0cm 0cm 0cm 0cm},clip]{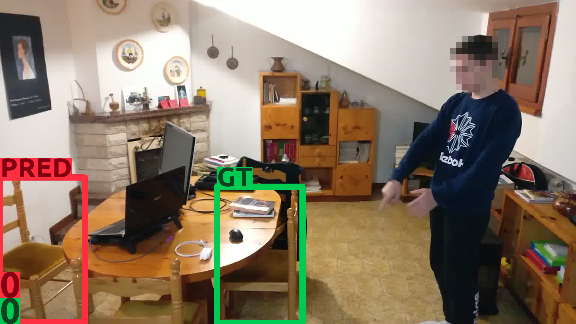}
    \end{subfigure}%
    \begin{subfigure}[t]{0.25\linewidth}
    	\includegraphics[width=\linewidth,trim={0cm 0cm 0cm 0cm},clip]{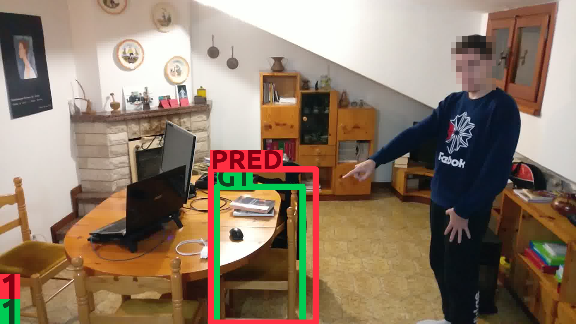}
    \end{subfigure}%
    \begin{subfigure}[t]{0.25\linewidth}
    	\includegraphics[width=\linewidth,trim={0cm 0cm 0cm 0cm},clip]{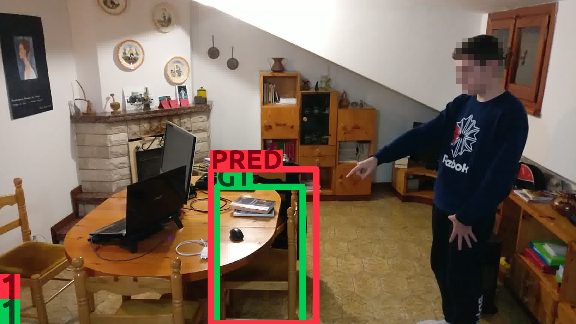}
    \end{subfigure}%
    \\
    \begin{subfigure}[t]{0.25\linewidth}
    	\includegraphics[width=\linewidth,trim={0cm 0cm 0cm 0cm},clip]{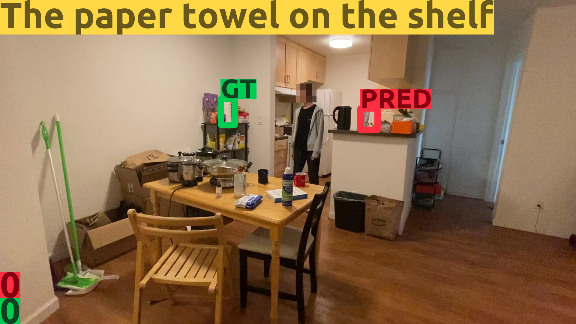}
    \end{subfigure}%
    \begin{subfigure}[t]{0.25\linewidth}
    	\includegraphics[width=\linewidth,trim={0cm 0cm 0cm 0cm},clip]{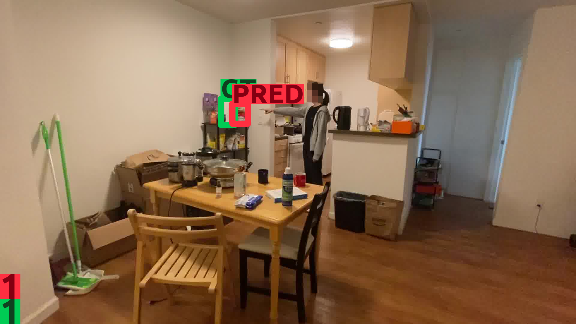}
    \end{subfigure}%
    \begin{subfigure}[t]{0.25\linewidth}
    	\includegraphics[width=\linewidth,trim={0cm 0cm 0cm 0cm},clip]{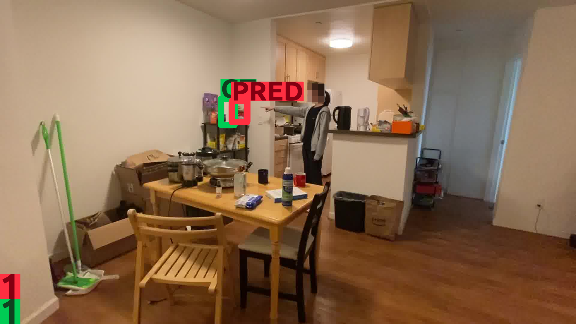}
    \end{subfigure}%
    \begin{subfigure}[t]{0.25\linewidth}
    	\includegraphics[width=\linewidth,trim={0cm 0cm 0cm 0cm},clip]{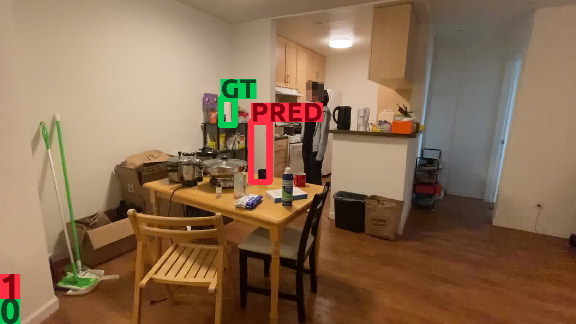}
    \end{subfigure}%
    \caption{\textbf{Qualitative results in Video \ac{eru} of the ConvLSTM model.} Each row represents four selected frames from one reference clip. Green/red boxes indicate the predicted/ground-truth reference targets. $0$ denotes non-canonical frame, and $1$ canonical frame.}
    \label{fig:video_eru_qual}
\end{figure*}

\paragraph{Results and Discussion}

\begin{figure}[t!]
	\centering
	\includegraphics[width=\linewidth,trim={0.8cm 0.1cm 0.8cm 2.1cm},clip]{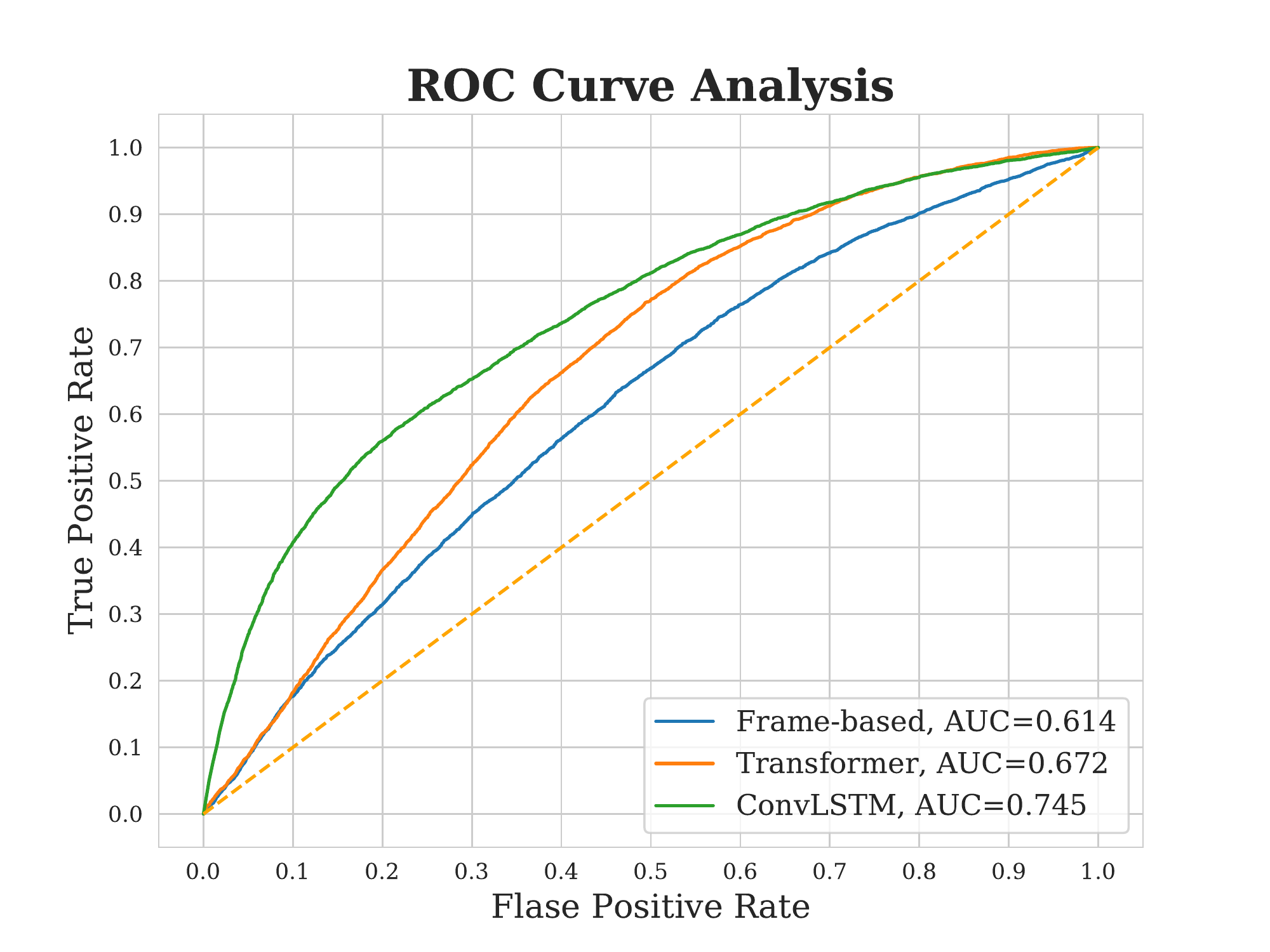}
	\caption{ROC Curve for canonical frame detection.}
	\label{fig:roc}
\end{figure}

\cref{tab:video_eru} shows quantitative results of predicting reference targets with the ground-truth canonical frames given a video. We observe that the frame-based method and the temporal optimization methods reach similar performance, comparable to the model that only trained on selected canonical frames (\ie, Ours$_{\text{Full}}$). This result indicates that the canonical frames can indeed provide sufficient language and gestural cues for clear reference purposes, and the temporal models may be distracted from non-canonical frames. This observation aligns with the settings of previous \ac{ref} tasks. Meanwhile, as shown in \cref{tab:frame_det,fig:roc}, temporal information can significantly improve the performance of canonical frame detection; both the \textit{ConvLSTM} and the \textit{Transformer} model outperform the \textit{Frame-based} method by a large margin. These results indicate the significance of distinguishing various stages of reference behaviors, \eg, initiation, canonical moment, and ending, for better efficacy in embodied reference understanding. \cref{fig:video_eru_qual} shows some qualitative results.

\begin{table}[ht!]
    \centering
    \caption{Canonical frame detection performance.}
    \begin{tabular}{lccc}
        \toprule
        Method      & Avg. Prec & Avg. Rec & Avg. F1 \\
        \midrule
        Frame-based &  31.9 & 37.7 & 34.5   \\
        Transformer &  35.1 & \textbf{44.2} & 39.1\\
        ConvLSTM    &  \textbf{57.0} & 37.9 & \textbf{45.4}\\
        \bottomrule
    \end{tabular}
    \label{tab:frame_det}
\end{table}

\setstretch{1}

\section{Conclusion and Future Work}

We present the novel problem of embodied reference understanding. Such a setting with both language and gestural cues is more natural for understanding human communication in our daily activities. To tackle this problem, we crowd-source the \textsl{YouRefIt} dataset and devise two benchmarks on images and videos. We further propose a multimodal framework and conduct extensive experiments with ablations. The experimental results provide strong empirical evidence that language and gestural coordination is critical for understanding human communication.

Our work initiates the research on embodied reference understanding and can be extended to many aspects. For example, the difficulty in resolving reference ambiguity within a single-round communication, even for humans, calls for studying embodied reference using multi-round dialogues. Human-robot interaction may benefit from referential behavior generation by considering scene contexts. We hope our work can inspire more future work on these promising directions, focusing on understanding human communication from multimodal (verbal/nonverbal) inputs.

\clearpage
{
\fontsize{9.8}{11.5}\selectfont
\balance{}
\bibliographystyle{ieee_fullname}
\bibliography{reference}
}

\end{document}